\documentclass[10pt,twocolumn,letterpaper]{article}

\usepackage{iccv}
\usepackage{times}
\usepackage{epsfig}
\usepackage{graphicx}
\usepackage{amsmath}
\usepackage{amssymb}
\usepackage[accsupp]{axessibility}

\usepackage{mathrsfs}
\usepackage{bm}
\usepackage{multirow}
\usepackage{xcolor}
\usepackage{bbding}
\usepackage[pagebackref=true,breaklinks=true,letterpaper=true,colorlinks,bookmarks=false]{hyperref}
\usepackage{appendix}

\usepackage[breaklinks=true,bookmarks=false]{hyperref}

\iccvfinalcopy 

\def\iccvPaperID{****} 

\ificcvfinal\fi

\begin{document}

\title{Synchronize Feature Extracting and Matching: A Single Branch Framework for 3D Object Tracking}

\author{Teli Ma$^{1,2}$\thanks{Equal Contribution}, Mengmeng Wang$^{2*}$, Jimin Xiao$^{3}$, Huifeng Wu$^{4}$, Yong Liu$^{2}$\thanks{Corresponding Author} \\
$^1$The Hong Kong University of Science and Technology, Guangzhou \quad 
$^2$Zhejiang University \quad \\
$^3$Xi'an Jiaotong-Liverpool University \quad 
$^4$Hangzhou Dianzi University \quad\\
{\tt\small tma184@connect.hkust-gz.edu.cn} \quad {\tt\small mengmengwang@zju.edu.cn} \\ {\tt\small jimin.xiao@xjtlu.edu.cn} \quad {\tt\small whf@hdu.edu.cn} \quad {\tt\small yongliu@iipc.zju.edu.cn}
}

\maketitle
\ificcvfinal\thispagestyle{empty}\fi

\begin{abstract}
   Siamese network has been a de facto benchmark framework for 3D LiDAR object tracking with a shared-parametric encoder extracting features from template and search region, respectively. This paradigm relies heavily on an additional matching network to model the cross-correlation/similarity of the template and search region. In this paper, we forsake the conventional Siamese paradigm and propose a novel single-branch framework, \textbf{SyncTrack}, synchronizing the feature extracting and matching to avoid forwarding encoder twice for template and search region as well as introducing extra parameters of matching network. The synchronization mechanism is based on the dynamic affinity of the Transformer, and an in-depth analysis of the relevance is provided theoretically. Moreover, based on the synchronization, we introduce a novel Attentive Points-Sampling strategy into the Transformer layers (APST), replacing the random/Farthest Points Sampling (FPS) method with sampling under the supervision of attentive relations between the template and search region. It implies connecting point-wise sampling with the feature learning, beneficial to aggregating more distinctive and geometric features for tracking with sparse points.
Extensive experiments on two benchmark datasets (KITTI and NuScenes) show that SyncTrack achieves state-of-the-art performance in real-time tracking.
\end{abstract}

\begin{figure}[]
\centering
\includegraphics[width=1.0\linewidth,trim={0cm 0cm 0cm 0cm}]{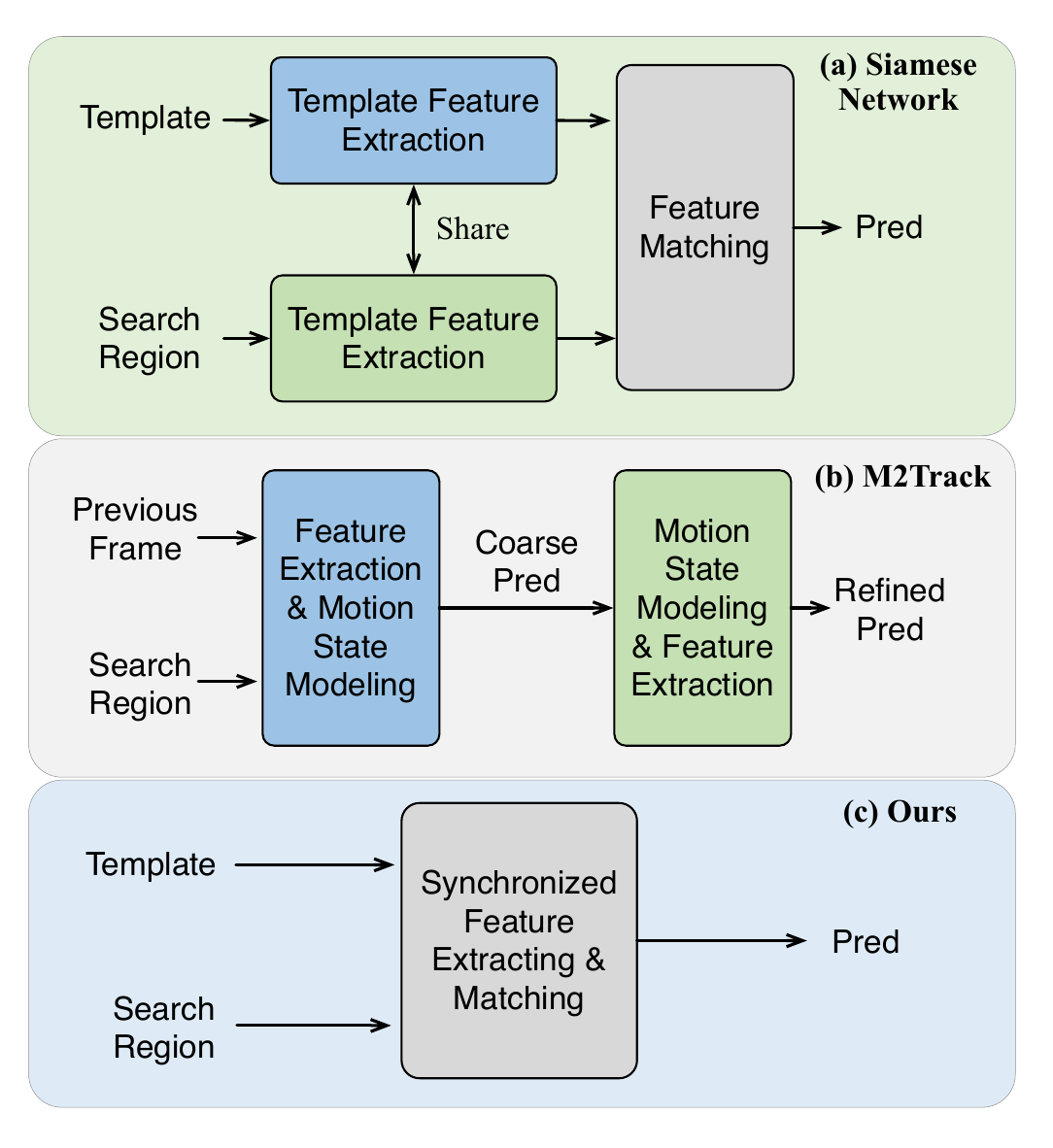}
\caption{The comparison of (a) Siamese network based trackers, (b) previous single-branch, two-stage framework M2Track~\cite{zheng2022beyond} with (c) our SyncTrack, which is a single-branch and single-stage framework.}
\label{fig:comparison}
\end{figure}

\section{Introduction}
\label{sec:intro}

With recent advances in autonomous driving, 3D vision tasks based on LiDAR are becoming increasingly popular in the visual community. Among these tasks, 3D LiDAR single object tracking (SOT) aims to track a specific target object in a 3D video with the knowledge of 3D bounding  box in the initial frame. This task meets numerous challenges, such as LiDAR point cloud sparsity, occlusions, and fast motions.

Most existing methods ~\cite{qi2020p2b, zheng2021box, hui20213d, giancola2019leveraging, shan2021ptt, zhou2022pttr, stnet, fang20203d, wang2021mlvsnet} of 3D SOT mainly adopt a Siamese-like backbone and incorporate an additional matching network to cope with the tracking challenges as shown in Fig.~\ref{fig:comparison}(a). Trackers based on the Siamese-like backbone separate  feature extraction of template and search region, forwarding the two kinds of features with shared model parameters, respectively. Subsequently, an extra matching network is introduced to fuse the extracted template and search region features to model the correlation or similarity between them. 
However, such a paradigm restricts the feature interaction to a  post-matching network, correlating the template and search region insufficiently with merely the high-level extracted features. 
In other words, the matching process posterior to the encoder is incapable of modeling the relations of multi-scale features intra-backbone.
Moreover, a standalone matching network results in extra model parameters and computational overheads, let alone the double forwarding process of the Siamese-backbone to extract the template and search region features. M2Track~\cite{zheng2022beyond} proposed a motion-centric paradigm to replace the Siamese-like structure, constructing a spatial-temporal point cloud to predict the motion. However, they still rely heavily on an additional motion transformation network, which requires extra training input, to integrate the extracted template features into the search region, and another two-stage refinement network is leveraged to ensure the performance as illustrated in Fig.~\ref{fig:comparison}(b). Based on aforementioned problems, we ask the question: \textit{Can feature extracting and matching be conducted simultaneously in a simple way?}

The answer is \textbf{Yes} and the solution is implied in the \textbf{dynamic global reasoning property} of Transformer~\cite{transformer, vit, gao2022mcmae}. Specifically, the affinity matrix of all tokens can be constructed dynamically via continuous computation of the \textit{key} and \textit{query} vectors in the attention mechanism. The spatial context is aggregated using affinity to attend features. Intuitively, this affinity matrix can intrinsically serve as the \textbf{\textit{matching matrix}} for intermediate feature interactions between the template and search region if we merge them into one input of the Transformer layers. Therefore, we propose a single-branch and single-stage framework equipped with a Transformer-based backbone instead of the conventional Siamese-like PointNet++~\cite{qi2017pointnet} backbone, as shown in Fig~\ref{fig:comparison}(c).
The framework is dubbed as SyncTrack, as the Transformer backbone synchronizes the feature extracting and matching process. The SyncTrack is composed of a simple backbone and prediction head, omitting the complex matching network design and motion state estimation, depending merely on point-wise features.

\begin{figure*}[]
\centering
\includegraphics[width=1.0\linewidth,trim={0cm 0cm 0cm 0cm}]{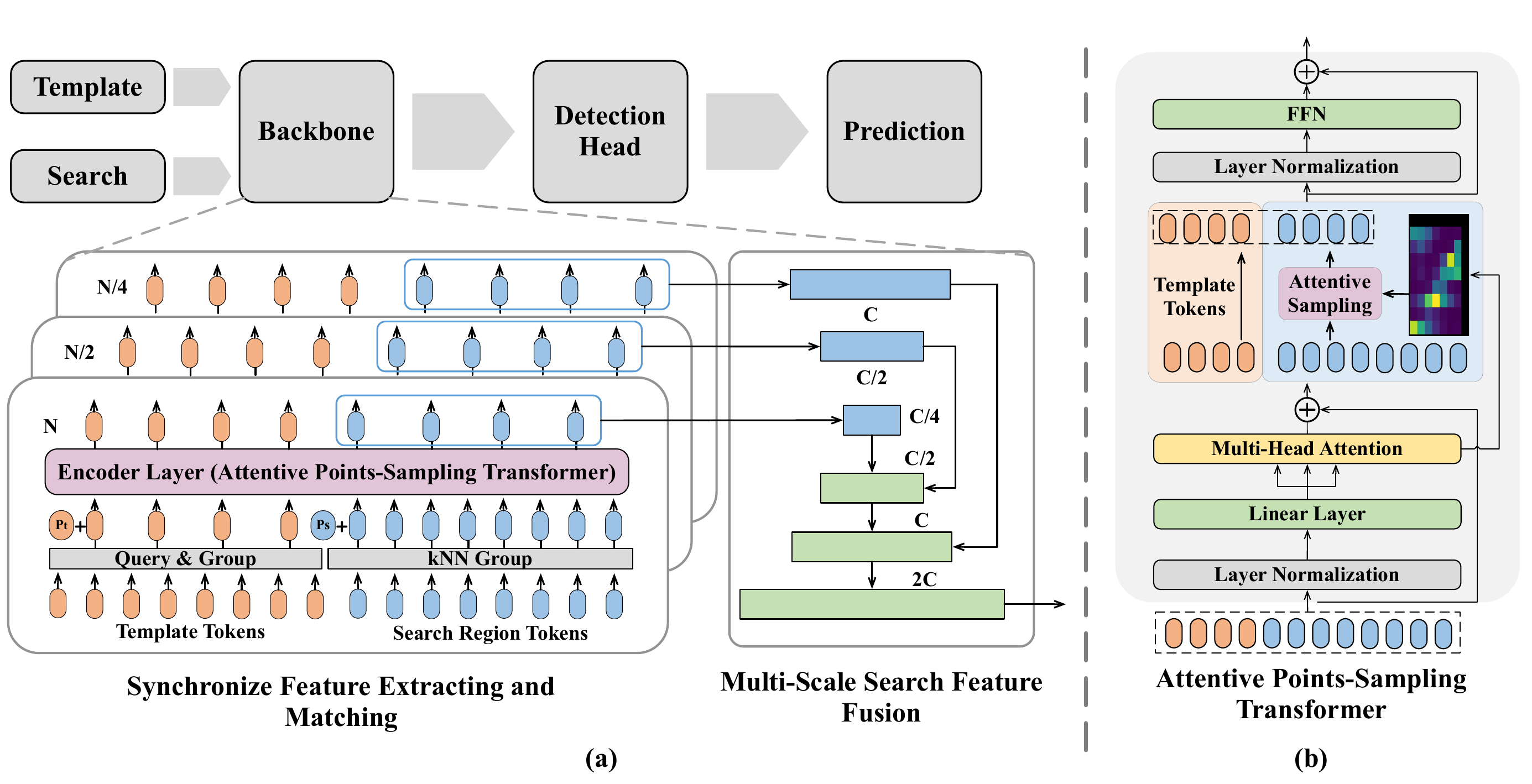}
\caption{\textbf{(a).} The overall framework of SyncTrack and specific illustration of the single-branch backbone, the template and search region input are concatenated for feature extraction and matching synchronously. \textbf{(b).} The structure of the Attentive Points-Sampling Transformer (APST), which samples search region points after multi-head attention.}
\label{fig:pipeline}
\end{figure*}

However, 3D point clouds have unique properties such as sparsity~\cite{zhuang2021perception, li2022coarse3d}, density variance, and implicit geometric features in data locality~\cite{mao2019interpolated}. For example, $51\%$ samples of KITTI~\cite{kitti} \textit{Car} category have less than 100 points~\cite{hui20213d}. These problems are further aggravated when point clouds are grouped and down-sampled to formulate multi-scale point-wise feature maps. The phenomena stresses the vitality of point clouds sampling, aiming to improve the point-wise perception efficiency with limited points.
PTTR~\cite{zhou2022pttr} proposed to sample the input point clouds before the backbone, utilizing the $L_2$ distance as the similarity metric for sampling. However, as down-sampling layer by layer is essential in tracking backbone for multi-scale feature fusion to strengthen representation, it is reasonable to consider the sampling strategy in the backbone.
Therefore, we propose the attentive sampling strategy based on the attention map between the template and search region, and equip each Transformer layer with our sampling module as shown in Fig.~\ref{fig:pipeline}(b). We name the Transformer containing Attentive Points-Sampling as APST. 
Specifically, the attentive response from template tokens to search region tokens is considered, as the positively respond tokens are more likely to be in the foreground and should be preserved for feature extracting. By contrast, as Fig.~\ref{fig:sampling} shows, random sampling easily falls into the perceptive confusion as selected points hardly contain geometric features due to randomness.

The main contributions of our paper can be summarized as follows:
\begin{itemize}
    \item We introduce a single-branch and single-stage framework for real-time 3D LiDAR SOT dubbed SyncTrack,  without Siamese-like forward propagation and a standalone matching network. We ingeniously leverage the dynamic affinity characteristic of the self-attention mechanism to synchronize the feature extracting and matching. A detailed analysis is provided to explain the synchronizing mechanism.
    \item We propose a novel APST to build the backbone, replacing the random/FPS\footnote{In this paper, we use FPS to refer to Farthest Points Sampling and fps to denote frames per second.} down-sampling of point clouds with attentive sampling to preserve more target-relevant points, and thus improving the perceptive capability of feature extracting.
    \item Extensive results show that our method has achieved new state-of-the-art performance on the KITTI and NuScenes datasets in real-time tracking, up to $2.8\%$ and $1.4\%$ on \textit{mean} results with a high-speed of around $45$ fps. Besides, SyncTrack exhibits good scalability in both width and depth.
\end{itemize}

\section{Related Works}
\subsection{2D Visual Tracking}
Advances in 2D video~\cite{zhou2021graph, liang2019peeking, liang2020garden} and object tracking approaches \cite{nam2016learning,held2016learning,wang2017large,li2018high,valmadre2017end,zhao2021deep} stimulate the development of the 3D tracking, and the methods are evolving all the time. Early 2D SOT methods mainly focus on the classifier design like structured output  SVM~\cite{wang2017large,hare2015struck}, correlation filter~\cite{henriques2014high,li2014scale}.
With the prevalence of deep learning, end-to-end trainable trackers emerged in the visual community. The Siamese-like structure~\cite{dong2019quadruplet,huang2019got,he2018twofold,li2018high,zhang2019deeper,xu2020siamfc,lu2020deep} based methods have been popular in the tracking field and have numerous variants.
SiamFC~\cite{bertinetto2016fully} is a pioneering work integrating feature correlation into a fully convolutional Siamese network for visual tracking. 
Subsequently, improvements like introducing detection components such as region proposal network~\cite{li2018high, li2019siamrpn++, fan2019siamese}, discriminating fore and back-ground~\cite{zhao2021adaptive}, anchor-free detecting~\cite{guo2020siamcar} \textit{etc.}.
Recently, vision transformers have been introduced into 2D SOT \cite{chen2021transformer,wang2021transformer,yan2021learning, xie2022correlation} to exploit the long-range modeling of the attention mechanism to fuse the features effectively. 
Moreover, the one-stream network trackers based on transformers are proposed in 2D tracking. The OSTrack~\cite{ostrack} proposes joint feature learning and relation modeling, masking a proportion of image patches to save computational overheads. SimTrack~\cite{chen2022backbone} concatenates the template and search input, improving the patch embedding method with a Foveal window strategy. 
Our work draws inspiration from the aforementioned trend in 2D vision. However, it is specifically tailored to address the challenges of the 3D Single Object Tracking (SOT) problem, taking into account the unique characteristics of point clouds. 
Furthermore, we analyze the success of the transformer-based single-branch backbone, attributing to the dynamic affinity of the attention mechanism, which enables the synchronization of feature extraction and matching processes.

\subsection{3D Visual Tracking}
In this paper, we only discuss LiDAR-based 3D object tracking. Till to now, almost all the 3D tracking methods~\cite{giancola2019leveraging,qi2020p2b,fang20203d,zarzar2019efficient,cui20213d,shan2021ptt,zhou2022pttr,zheng2021box,hui20213d, stnet, wang2023correlation} are based on the Siamese structure. 
The pioneering work in this field is the SC3D~\cite{giancola2019leveraging}, which initially defines the task. SC3D employs cosine similarity to measure the resemblance between template and search region features, and incorporates shape completion during training to enhance appearance refinement.
 Trackers following SC3D make advancements from two perspectives. Firstly, they enhance the matching network~\cite{qi2020p2b,zheng2021box,hui20213d,zhou2022pttr,cui20213d,wang2021mlvsnet,zheng2022beyond}. 
 For instance, MLVSNet \cite{wang2021mlvsnet} uses the CBAM module \cite{woo2018cbam} to enhance the vote cluster features with both channel and spatial attention. 
 STNet~\cite{stnet} employs cross- and self-attention modules to enhance the interaction between the extracted template and search region features, boosting their feature-level integration.
 M2Track introduces a motion-centric paradigm and motion estimation module to correlate the template and search features instead of appearance matching. All these methods depend on a standalone module to match the features.
Secondly, the trackers \cite{qi2020p2b,fang20203d,hui20213d,wang2021mlvsnet,shan2021ptt} have attempted to improve the prediction head part. P2B \cite{qi2020p2b} employs Hough Voting to predict the target location, and ~\cite{zheng2021box, shan2021ptt,wang2021mlvsnet} all follow the voting strategy to make predictions. 3D-SiamRPN \cite{fang20203d} uses an RPN head to predict the final results. 
LTTR \cite{fang20203d} and V2B \cite{hui20213d} use center-based regressions to predict several object properties. 
However, few efforts are made in exploring the encoder/backbone of trackers as PointNet++~\cite{qi2017pointnet} is the feature extractor by default.
In this study, we focus on the backbone design and incorporate the matching process directly into the backbone, significantly streamlining the tracker network.

\section{Method}
In this section, we first define the 3D SOT task in Sec.~\ref{sec:definition}. Then, a detailed introduction of the single-branch framework is included in the Sec.~\ref{sec:single-branch}. 
Based on the single-branch framework, how the feature extracting and matching are synchronized is elaborated in (Sec.~\ref{sec:sync}). 
Moreover, we propose the Attentive Points-Sampling Transformer to build the single-branch backbone and sampling search region tokens with a strategy of attentive sampling, as shown in Sec.~\ref{sec:apst}. The decoder head and losses are described in Sec.~\ref{sec:decoder}.

\subsection{Problem Definition}
\label{sec:definition}
In the configuration of 3D LiDAR single object detection (SOT) task,  
the 3D bounding box (BBox) is defined as $(x, y, z, w, l, h, \theta) \in \mathbb{R}^7$ , where the $(x, y, z)$ represents the coordinate center of the BBox and $(w, l, h), \theta$ stand for the BBox size and heading angle (the rotation around the \textit{up-axis}) respectively. Generally, the BBox size is assumed to be fixed by default even when the target object is non-rigid, thus minimizing the dimensions of BBox from $\mathbb{R}^7$ to $\mathbb{R}^4$. 
Given a sequence of temporally-connected point clouds $\mathcal{\{P}_i\}_{i=1}^T$ ($T$ is the number of points in each frame) and a initial BBox $\mathcal{B}_1$ of the target, the goal of SOT is to localize the target BBoxes $\mathcal{\{B}_i\}_{i=2}^T$ in all frames online. Following the previous manner, a template point cloud $\mathcal{P}^t = \{p_i^t\}_{i=1}^{N_{t}}$ and a search region  $\mathcal{P}^s = \{p_i^s\}_{i=1}^{N_{s}}$ are generated, where $N_t$ and $N_s$ are number of template and search region points. The template $P^t$ is generated by cropping and centering the target in the initial frame based on the initial BBox.

\subsection{Single-Branch Structure}
\label{sec:single-branch}
We propose to replace the conventional \textit{Siamese-like backbone} paradigm with a sole backbone, eliminating the double forward process of Siamese structure.
Therefore, template and search region seeds are concatenated to do forward propagation. 
The Transformers' property of long-range relation-modeling is the intrinsic merit to tackle the concatenated template and search region seeds. Based on that, we leverage the self-attention modules to build the single-branch backbone as shown in Fig.~\ref{fig:pipeline}(a). 
In our approach, we utilize a Query \& Group module to sample the template seeds, denoted as $\mathcal{P}^t \in \mathbb{R}^{N_t\times C}$, using the FPS method before jointly forwarding. This module also groups the k-nearest points to aggregate features. However, when it comes to the search region point cloud, represented as $\mathcal{P}^s \in \mathbb{R}^{N_s\times C}$, we solely employ the Group module to aggregate neighbor information without reducing the number of search points.
Subsequently, the template and search seeds are concatenated, incorporating a joint parametric positional embedding for the localization of tokens. This process is as follows:
\begin{equation}
\begin{aligned}
  \mathcal{T}^t =&  {\rm kNN(FPS}(\mathcal{P}^t)), \mathcal{P}^t \in \mathbb{R}^{N'_t \times C'}, N'_t < N_t, \\ 
   \mathcal{T}^s &= {\rm kNN}(\mathcal{P}^s), \mathcal{P}^s \in \mathbb{R}^{N_s \times C'}, \\
   \mathcal{T}^{ts} &=  [\mathcal{T}^t; \mathcal{T}^s] + \mathbf{p_{e}}, \mathcal{T}^{ts} \in \mathbb{R}^{(N_t+N_s) \times C'}.
\end{aligned}
\end{equation}
Afterward, linear layers are leveraged to project the input tokens into \textit{query}, \textit{key}, and \textit{value} latent and the head-wise joint attention map is calculated to model the intra- \& inter-relations of template and search tokens.

\begin{equation}
\begin{aligned}
    Q^{ts}, K^{ts}, V^{ts} &= W_q(\mathcal{T}^{ts}),  W_k(\mathcal{T}^{ts}),  W_v(\mathcal{T}^{ts}), \\
    \mathcal{A}_m^{ts} =& {\rm Softmax}\frac{Q_m^{ts} (K_m^{ts})^{\top}}{\sqrt{C'/M}}.
\end{aligned}
\label{eq:joint_attn}
\end{equation}
Based on the joint head-wise attention map $\mathcal{A}_m^{ts}$ of template and search tokens, the features extracted by multi-head attention are:
\begin{equation}
    \mathcal{T'}^{ts} = [\mathcal{A}_1^{ts}V^{ts}_1, \mathcal{A}_2^{ts}V^{ts}_2,\dots,\mathcal{A}_M^{ts}V^{ts}_M]\mathcal{W} + \mathcal{T}^{ts},
\end{equation}
where $M$ is the number of the attention heads and $\mathcal{W}$ is the weight of a MLP.

\subsection{Synchronize Feature Extracting and Matching}
\label{sec:sync}
We illustrate how the single-branch structure can synchronize the process of feature extracting and matching with a simple backbone. The formula of joint attention $\mathcal{A}_m^{ts}$ in Eq.~\ref{eq:joint_attn} can be expanded as ($\sigma$ represents \textit{Softmax} operation):
\begin{equation}
\begin{aligned}
[\mathcal{A}_m^{t}; \mathcal{A}_m^{s}] &= \bm{\sigma} \frac{[Q_m^{t}; Q_m^{s}] [K_m^{t}; K_m^{s}]^{\top}}{\sqrt{C'/M}} \\
=[\bm{\sigma}\frac{Q_m^{t}(K_m^{t})^\top}{\sqrt{C'/M}}, &\bm{\sigma}\frac{Q_m^{t}(K_m^{s})^\top}{\sqrt{C'/M}}; \bm{\sigma}\frac{Q_m^{s}(K_m^{t})^\top}{\sqrt{C'/M}}, \bm{\sigma}\frac{Q_m^{s}(K_m^{s})^\top}{\sqrt{C'/M}}].
\end{aligned}
\label{eq:attnmap}
\end{equation}
The new extracted search region features can be obtained based on Eq~\ref{eq:attnmap} as:   
\begin{equation}
\begin{aligned} 
\mathcal{T'}_m^{s} = \bm{\sigma}\frac{Q_m^{s}(K_m^{t})^\top}{\sqrt{C'/M}}&V_m^{t} + \bm{\sigma}\frac{Q_m^{s}(K_m^{s})^\top}{\sqrt{C'/M}}V_m^{s}, \\
 [V_m^t; V_m^s] &= V_m^{ts},
\end{aligned}
\end{equation}
depending on the projected features $[V_m^t, V_m^s]$ of both the template and search region from last layer. The attention queried from search region to template, $\bm{\sigma}\frac{Q_m^{s}(K_m^{t})^\top}{\sqrt{C'/M}}$,  is the \textbf{\textit{matching matrix}} that guides to aggregate highly-relevant template features. 
Moreover, the \textbf{\textit{matching matrix}} is dynamic, which is determined by the changing \textit{query} and \textit{key} latent of search and template features as:
\begin{equation}
\begin{aligned}
   &\bm{\sigma}\frac{Q_m^{s}(K_m^{t})^\top}{\sqrt{C'/M}} =  \bm{\sigma}\frac{W_{q,m}^s \mathcal{T}_m^{s} (\mathcal{T}_m^{t})^\top (W_{k,m}^{t})^\top}{\sqrt{C'/M}}, \\
   &i.e.~~~~~~~ \bm{\sigma}\frac{Q_m^{s}(K_m^{t})^\top}{\sqrt{C'/M}}\propto  [\mathcal{T}_m^{s}; \mathcal{T}_m^{t}].
\end{aligned}
\label{eq:dynamic}
\end{equation}
To conclude, the synchronization attributes to the dynamic mechanism of \textbf{\textit{matching matrix}}, continuously adapting the matching relations according to the extracted features of template and search region. 

\noindent\textbf{Comparisons with Siamese Network.~}
Previous Siamese-like trackers can be summarized as the paradigm of \textit{Extracting then Matching}. If we name the feature extracting and matching as $\phi$ and $\delta$, then the objective of this paradigm can be concluded as ${max}(\delta(\phi(\mathcal{P}^t), \phi(\mathcal{P}^s))$ for simplicity. It means the model training is to shorten the distances between correlated parts of the template and search region based on the extracted features from the backbone. 
However, compared to our approach, this matching mechanism is relatively \textit{static} since it occurs only after feature extraction, resulting in inadequate modeling of inter-backbone relations. On the other hand, our single-branch framework facilitates dynamic interaction between the search region seeds and the template across all layers of the backbone. This allows for comprehensive learning of relations, encompassing both the local representation from early layers and the global representation from later ones.

\begin{figure}[]
\centering
\includegraphics[width=1.0\linewidth,trim={0cm 0cm 0cm 0cm}]{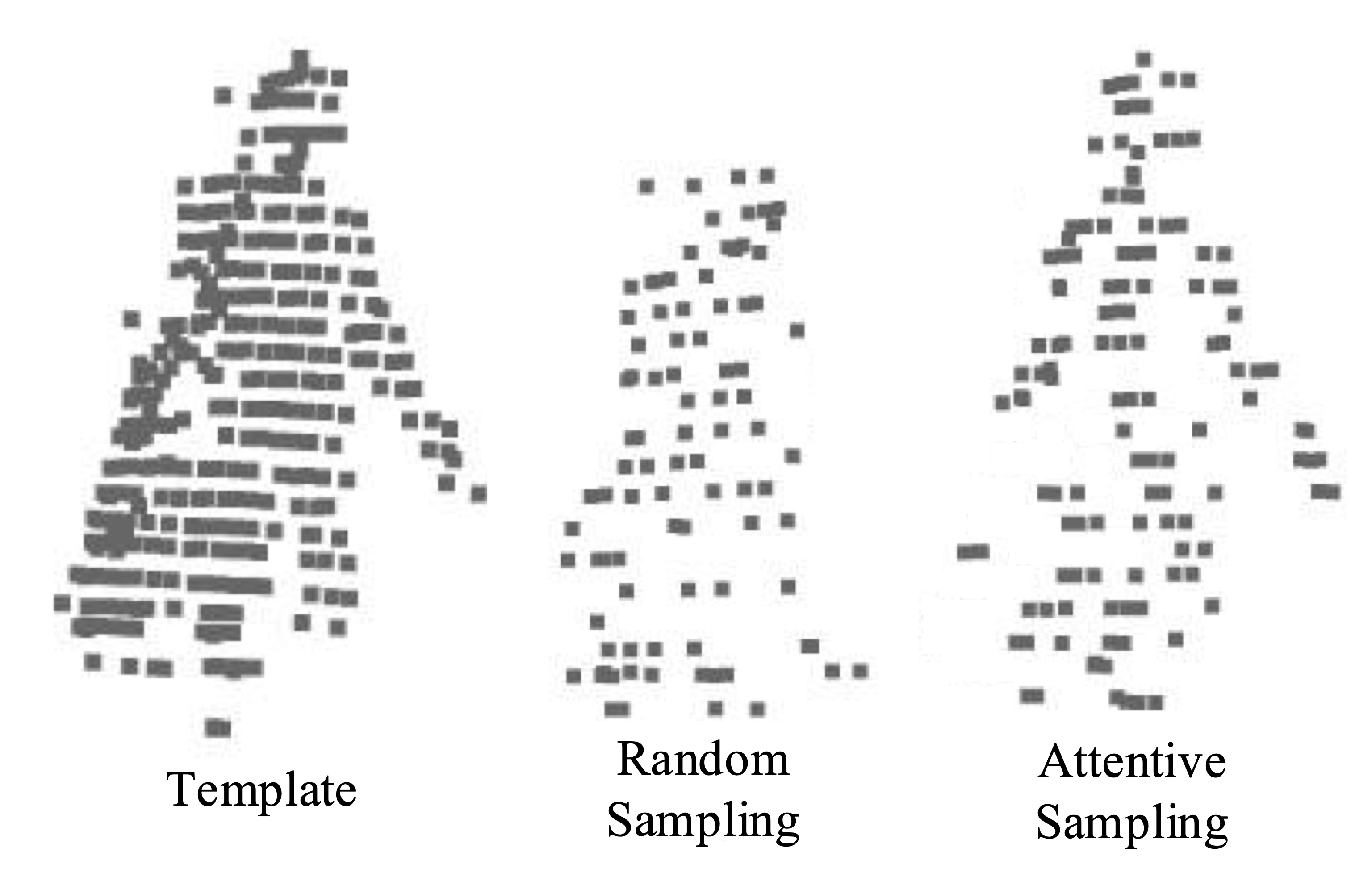}
\caption{\textbf{Comparisons of random sampling with our attentive sampling in point-wise down-sampling.} We find that our attentive sampling approach excels in selecting more representative points, including those that contribute to the formation of the head, arms, and legs, which contain distinctive geometric features. In contrast, random sampling include more points that are far away from the target (we have removed such points for better visualization). }
\label{fig:sampling}
\end{figure}

\subsection{Attentive Points-Sampling Transformer}
\label{sec:apst}
Based on the synchronizing mechanism of feature extracting and matching, we propose to replace the Point-wise Transformer~\cite{zhao2021point} with a novel Attentive Points-Sampling Transformer (APST). APST is proposed based on the observation that backbones utilized in previous 3D LiDAR trackers always adopt a non-parametric strategy of down-sampling the search region features/tokens with farthest point sampling (FPS) or randomly sampling to reduce the points as introduced in PointNet++~\cite{qi2017pointnet}.
Nevertheless, this non-parametric sampling method lacks learnability and controllability since no parameters associated with the sampling process are updated during model training.
Consequently, the final performance is compromised due to the significant influence of the points-center in LiDAR, which determines the effectiveness of extracting features around the foreground, as shown in Fig.~\ref{fig:sampling}.

Therefore, we introduce the APST, which involves the selection of points based on attentive relations between the tokens of the template and search region as illustrated in sub-figure (b) of Fig.~\ref{fig:pipeline}. The attentive response of template tokens reacting to search region tokens is considered as positively responding search tokens are more likely to be in the foreground. In that case, we segment the attention map and separate the $\mathcal{A}_{m}^{t\rightarrow s}=\bm{\sigma}\frac{Q_m^{t}(K_m^{s})^\top}{\sqrt{C'/M}}$ from the Eq.~\ref{eq:attnmap}, averaging along the dimension of template-tokens attention among all transformer heads to acquire the response scores. To ensure the maximum response of search tokens to template ones, a group of search region indexes, named $\Omega_s$, is dynamically updated as:
\begin{equation}
    \Omega_s^{*} = \mathop{\arg\max}_{\Omega_s} \frac{1}{N_t}\frac{1}{M}\sum\limits_{t=0}\limits^{N_t}\sum\limits_{m=0}\limits^{M}\frac{Q_m^{t}}{\sqrt{C'/M}}(K_m^{s})^\top\bigg|_{s\in \Omega_s},
\end{equation}
Based on the optimal solution $\Omega_{s}^{*}$, the search region tokens are sampled to decrease the number of tokens after multi-head self-attention and concatenate with template ones. Note that the template tokens are sampled with the FPS method as introduced in Sec.~\ref{sec:single-branch}, and only the search region tokens are sampled with the attentive sampling method.

The dynamic-affinity property of the Transformer suggests that the attention map is influenced by the latent tokens projected through learnable linear layers. Hence, it is reasonable to assert that sampling tokens guided by the attention map is linked to model learning, as the attention map is generated based on updated parameters. As a result, this type of token/point sampling proves advantageous in providing prior knowledge for centroid selection and enhancing the efficient aggregation of representational information, as shown in Fig.~\ref{fig:sampling}.

\subsection{Decoder and Losses}
\label{sec:decoder}
With the encoded point-wise features of various scales, a multi-scale feature fusion module is adopted to fuse these features and output the features with the same origin input size, feeding into the decoder part for final predictions.
Following the V2B~\cite{hui20213d}, the features are voxelized as a volumetric representation and 3D convolutions are utilized on the encoded features. Afterward, the BEV feature maps are acquired by pooling on the \textit{z}-axis for regression. Focal loss and L1 loss are leveraged for classification and BBox center offset and rotation regression, respectively. A detailed introduction is presented in the supplementary.

\section{Experiments}

\subsection{Experiment Setups}
\noindent\textbf{Implementation Details.~}
We set the number of input points as $N_t=512$ and $N_s=1024$ for template and search regions by randomly duplicating and discarding, respectively. The encoder backbone is merely consisted of three layers with Attentive Points-Sampling Transformers, and the number of both template and search region points output by each layer are 256, 128 and 64. Moreover, each layer's feature dimensions in the encoder are 32, 64 and 128, respectively, whereas the final features for the prediction head are with 32 channels. Note that the heads for all APST are 2 by default. In the voxelization process, the region $[(x_{min},x_{max}),(y_{min},y_{max}),(z_{min},z_{max})]$ is defined as [(-5.6,5.6),(-3.6,3.6),(-2.4,2.4)] to contain most target points. The voxel size $(v_x,v_y,v_z)$ is set to (0.3,0.3,0.3). For the detection head, four decomposed 3D (stride of 2,1,2,1 along the \textit{z}-axis) and 2D convolution blocks (stride of 2,1,1,2) are leveraged to strengthen the feature aggregation.

\noindent\textbf{Training and Testing.~} 
We train the model for 40 epochs with a batch size of 64. The Adam optimizer~\cite{adam} is adopted with the initial learning rate of 0.001 and reduced by 5 every 10 epochs(every 2 epochs for nuScenes). The classification loss has a weight $\lambda_{cls}$ of 1 and the regression loss has a weight $\lambda_{reg}$ of 1.

\noindent\textbf{Evaluation Metrics.~} 
Following the previous methods~\cite{qi2020p2b,zheng2021box}, we measure the \textit{Success} and \textit{Precision} of the tracker. To be specific, \textit{Success} is defined as the IoU between predicted boxes and the ground truth, 
and \textit{Precision} measures the AUC (Area Under Curve) of the distance between predicted and ground truth boxes within the range of [0,2] meters.

\begin{table*}[htbp]
\centering
\small
\caption{\textit{Sucess}/\textit{Precision} comparisons among our SyncTrack and the state-of-the-art methods on the \textbf{KITTI} datasets. Mean shows frame-level averaging results. \textbf{Bold} and \underline{underline} denote the best
performance and the second-best performance, respectively. Improvements over previous state-of-the-arts are shown in \textit{Italic} and color.}
\label{tab:kitti}
 \resizebox{\textwidth}{!}{
\begin{tabular}{c|cc|cc|cc|cc|cc}
\hline
 & \multicolumn{2}{c|}{Car (6424)}  & \multicolumn{2}{c|}{Cyclist (308)}               & \multicolumn{2}{c|}{Van (1248)}  & \multicolumn{2}{c|}{Pedestrian (6088)}          & \multicolumn{2}{c}{Mean (14068)}                                            \\ \cline{2-11} 
\multirow{-2}{*}{Methods} & Success & Precision & Success & Precision  & Success & Precision & Success & Precision & Success  & Precision  \\ \hline \hline
\multicolumn{11}{c}{\textit{Siamese Network}}\\ \hline
SC3D \cite{giancola2019leveraging} & 41.3 &57.9 & 41.5 &70.4 &40.4 &47.0 &18.2 &37.8 & 31.2 &48.5 \\
SC3D-RPN\cite{zarzar2019efficient} & 36.3 &51.0  & 43.0 &81.4 & - &- & 17.9 &47.8   & - &-  \\
P2B \cite{qi2020p2b} & 56.2 &72.8 & 32.1 &44.7 & 40.8 &48.4 & 28.7 &49.6 & 42.4 &60.0 \\
MLVSNet \cite{wang2021mlvsnet} & 56.0 &74.0 & 34.3 &44.5 & 52.0 &61.4 & 34.1 &61.1 & 45.7 &66.6 \\
3DSiamRPN \cite{fang20203d} & 58.2 &76.2 & 36.1 &49.0 & 45.6 &52.8 & 35.2 &56.2 & 46.6 &64.9  \\
LTTR \cite{cui20213d} & 65.0 &77.1 & 66.2 &89.9  & 35.8 &45.6 & 33.2 &56.8 & 48.7 &65.8        \\
PTT \cite{shan2021ptt} & 67.8 &81.8 & 37.2 &47.3 & 43.6 &52.5 & 44.9 &72.0 & 55.1 &74.2        \\
BAT \cite{zheng2021box} & 60.5 &77.7  & 33.7 &45.4 & 52.4 &67.0 & 42.1 &70.1  & 51.2 &72.8     \\
V2B \cite{hui20213d} & 70.5 &81.3 & 40.8 &49.7  & 50.1 &58.0 & 48.3 &73.5 & 58.4 &75.2 \\
PTTR  \cite{zhou2022pttr} & 65.2 &77.4 & 65.1 &90.5 & 52.5 &61.8 & \underline{50.9} &\textbf{81.6}  & 58.4 &77.8 \\ 

STNet~\cite{stnet} &\underline{72.1} &\underline{84.0}  &\textbf{73.5} &\underline{93.7} &\underline{58.0} &\textbf{70.6} &49.9 &77.2 &\underline{61.3} &\underline{80.1}\\ \hline
\textbf{SyncTrack} & \textbf{73.3} &\textbf{85.0} & \underline{73.1} &\textbf{93.8}     & \textbf{60.3} &\underline{70.0} & \textbf{54.7} &\underline{80.5}   & \textbf{64.1} &\textbf{81.9} \\
improvement               & {\color[HTML]{CB0000} \textit{+1.2}} &{\color[HTML]{CB0000} \textit{+1.0}} & {\color[HTML]{009901} \textit{-0.4}} &{\color[HTML]{CB0000} \textit{+0.1}} & {\color[HTML]{CB0000} \textit{+2.3}} &{\color[HTML]{009901} \textit{-0.6}} & {\color[HTML]{CB0000} \textit{+3.8}} &{\color[HTML]{009901} \textit{-1.1}} & {\color[HTML]{CB0000} \textit{+2.8}} &{\color[HTML]{CB0000} \textit{+1.8}} \\
\hline
\multicolumn{11}{c}{\textit{Single Branch Network}}\\ \hline
M2Track~\cite{zheng2022beyond} & \underline{65.5} &\underline{80.8}   & \textbf{73.2} &\underline{93.5} & \underline{53.8} &\textbf{70.7}  &\textbf{61.5} &\textbf{88.2}   & \underline{62.9} &\textbf{83.4} \\ 
\textbf{SyncTrack} & \textbf{73.3} &\textbf{85.0} & \underline{73.1} &\textbf{93.8}     & \textbf{60.3} &\underline{70.0} & \underline{54.7} &\underline{80.5}   & \textbf{64.1} &\underline{81.9}                        \\
improvement               & {\color[HTML]{CB0000} \textit{+7.8}} &{\color[HTML]{CB0000} \textit{+4.2}} & {\color[HTML]{009901} \textit{-0.1}} &{\color[HTML]{CB0000} \textit{+0.3}} & {\color[HTML]{CB0000} \textit{+2.3}} &{\color[HTML]{009901} \textit{-0.7}} & {\color[HTML]{009901} \textit{-6.8}} &{\color[HTML]{009901} \textit{-7.7}} & {\color[HTML]{CB0000} \textit{+1.2}} &{\color[HTML]{009901} \textit{-1.5}} \\ \hline
\end{tabular}
}
\end{table*}

\begin{table*}[htbp]
\centering
\caption{Comparison among SyncTrack and the state-of-the-art methods on the \textbf{nuScenes} datasets. Mean shows the average result weighed by frame numbers. \textbf{Bold} and \underline{underline} denote the best
performance and the second-best performance, respectively. Improvements over previous state-of-the-arts are shown in \textit{Italic} and color.}
\label{tab:nuscene}
 \resizebox{\textwidth}{!}{
\begin{tabular}{c|cc|cc|cc|cc|cc}
\hline
 & \multicolumn{2}{c|}{Car (15578)}  & \multicolumn{2}{c|}{Bicycle (501)}               & \multicolumn{2}{c|}{Truck (3710)}  & \multicolumn{2}{c|}{Pedestrian (8019)}          & \multicolumn{2}{c}{Mean (27808)}                                            \\ \cline{2-11} 
\multirow{-2}{*}{Methods} & Success & Precision & Success & Precision  & Success & Precision & Success & Precision & Success  & Precision  \\ \hline \hline
SC3D \cite{giancola2019leveraging}  & 24.5  & 25.9  & 16.6 & 18.8  & 32.5    & 30.6  & 13.8                                 & 14.7                                 & 22.3                                 & 23.2                                 \\
P2B \cite{qi2020p2b}                        & 32.8                                 & 35.2                           & 19.7                                 & 26.6                                  & 16.2                                 & 11.1                                 & 19.2                                 & 26.6                                 & 26.4                                 & 29.3                                 \\
BAT \cite{zheng2021box}                        & 26.5                                 & 28.8                                 & 17.8                                 & 22.8                                  & 16.5                                 & 10.6                                 & 19.4                                 & \textbf{28.2}                           & 23.0                                 & 27.9                                 \\
V2B \cite{hui20213d}                        & 32.9                           & 34.5                                 & 20.3                          & 27.5                           & 28.7                                 & 23.8                                 & \underline{20.1}                           & 27.4                                 & 28.4                          & 30.9                           \\ 
STNet~\cite{stnet} &\underline{35.7} &\underline{37.2} &22.3 &29.3 &\underline{33.5} &\underline{32.4} &\underline{20.1} &27.8 &\underline{30.7} &\underline{33.7} \\
M2Track~\cite{zheng2022beyond} &31.4 &33.9 &\underline{22.6} &\underline{29.8} &30.1 &28.8 &\textbf{20.7} &\underline{28.0} &28.0 &31.4 \\
\hline
\textbf{SyncTrack}          & \textbf{36.7}                        & \textbf{38.1}                        & \textbf{23.8}                        & \textbf{30.4}                         & \textbf{39.4}                        & \textbf{38.6}                        & 19.1                        & 27.8                        & \textbf{31.8}                        & \textbf{35.1}                        \\
improvement               & {\color[HTML]{CB0000} \textit{+1.0}} & {\color[HTML]{CB0000} \textit{+0.9}} & {\color[HTML]{CB0000} \textit{+1.2}} & {\color[HTML]{CB0000} \textit{+0.6}} & {\color[HTML]{CB0000} \textit{+5.9}} & {\color[HTML]{CB0000} \textit{+6.2}} & {\color[HTML]{009901} \textit{-1.6}} & {\color[HTML]{009901} \textit{-0.4}} & {\color[HTML]{CB0000} \textit{+1.1}} & {\color[HTML]{CB0000} \textit{+1.4}} \\ \hline
\end{tabular}
}
\end{table*}

\subsection{Comparison with State-of-the-Art Trackers}
\noindent\textbf{Results on KITTI.~}
KITTI~\cite{kitti} is one of the most popular datasets used in mobile robotics and autonomous driving. The tracking benchmark of KITTI consists of 21 training sequences and 29 test sequences. Following the previous methods~\cite{giancola2019leveraging,zheng2022beyond,zheng2021box}, we split the training sequences into train/val/test splits due to the inaccessibility of the testing labels, scenes 0-16 for training, scenes 7-18 for validation and scenes 19-20 for testing, respectively.

We compare the SyncTrack with other state-of-the-art methods from the pioneering SC3D~\cite{giancola2019leveraging} to the most recent siamese network STNet~\cite{stnet} and single branch network M2Track~\cite{zheng2022beyond},  as shown in Table~\ref{tab:kitti}. We separate the trackers into \textit{Siamese Network} and \textit{Single Branch Network} categories to compare with our proposed SyncTrack. 
Compared with siamese structured trackers, the SyncTrack achieves the best results on rigid and non-rigid object tracking, outperforming current tracking methods based on siamese networks on most specific categories and the overall mean results. 
STNet~\cite{stnet} is the state-of-the-art siamese-based tracking method with self-attention and cross-attention modules to match the template and search-region features.
Our SyncTrack outperforms the STNet by a relatively large margin on \textit{Car}, \textit{Van} and \textit{Pedestrian} categories under the evaluation of \textit{Success} metric. Also, SyncTrack surpasses the previous best \textit{Mean} results by $2.8\%$ and $1.8\%$ on \textit{Success} and \textit{Precision} metrics, respectively.

We also compare the SyncTrack with the only current single-branch tracker M2Track~\cite{zheng2022beyond}. Our SyncTrack outperforms M2Track by a large margin,  up to $7.8\%$ in \textit{Success} in the category of \textit{Car} whereas the M2Track is better in the category of \textit{Pedestrian}. However, for the comprehensive evaluation, the \textit{Mean} performance of all frames, our SyncTrack outperforms the M2Track by $1.2\%$ on the \textit{Success} metric.

\noindent\textbf{Results on nuScenes.~}
The nuScenes dataset~\cite{nuscenes} contains 1000 driving scenes collected from Boston and Singapore with a diverse and exciting set of driving maneuvers, traffic situations, and unexpected behaviors. In the configurations of LiDAR-based tracking methods, the train/val/test sets make up 700/150/150 of the whole 1000 scenes, respectively.  Officially, the train set is evenly split into 'train track' and 'train detect' to remedy overfitting. Following \cite{hui20213d}, we train our model with “train track” split and test it on the val set.

Note that nuScenes dataset only annotates keyframes and provides official interpolated results for the remaining frames, so there are two configurations for this dataset. The first is from \cite{zheng2021box, zheng2022beyond} which trains and tests both only on the keyframes. The second one is from \cite{hui20213d, stnet}, which trains and tests on all the frames. The results based on these two configurations are different. 
We believe that the motion in key frames is substantial, which does not conform to the practical applications. Therefore, we train and test all the frames in this paper. We train the previous methods on the nuScenes by ourselves using their official codes to compare with our SyncTrack as many results are missing or only reported by testing nuScenes test-split with a pre-trained KITTI model.

As shown in Table~\ref{tab:nuscene}, SyncTrack performs significantly better than other trackers on the mean results of four categories. Specifically, the SyncTrack yields the best results on both metrics and on most categories except the \textit{Pedestrian}, which is lower than M2Track~\cite{zheng2022beyond} and BAT~\cite{zheng2021box} by a minor margin ($1.6\%$ and $0.4\%$). 
However, in the \textit{Truck} category, our SyncTrack outperforms state-of-the-art by a large margin, up to $5.9\%$ and $6.2\%$ on \textit{Success} and \textit{Precision} respectively.

\noindent\textbf{Computational Cost Comparison.~}
We analyze the computational overheads and inference speed of SyncTrack and compare it with other trackers in Table~\ref{tab:runtime}. The reported results are tested by ourselves with official codebases with a single TITAN RTX GPU on \textit{Car} category of KITTI. It can be seen that SyncTrack achieves the best \textit{Success} performance with the lowest computational complexity (2.51 G). Compared with the most current Siamese-based tracker STNet~\cite{stnet} and single-branch tracker M2Track, our SyncTrack has the fewest number of parameters and fastest inference speed, satisfying the demand of real-time tracking.
\begin{table}[htbp]
\centering
\small
\caption{The computational cost of different trackers.}
\label{tab:runtime}
\begin{tabular}{c|cccc}
\hline
Methods  & Parameters & FLOPs   & FPS  &Success \\ \hline
SC3D~\cite{giancola2019leveraging}    & 6.45 M     & 20.07 G & 6    &41.3 \\
P2B~\cite{qi2020p2b}     & 1.34 M     & 4.28 G  & 48     &56.2\\
BAT~\cite{zheng2021box}     & 1.47 M     & 5.53 G  & 54   &60.5  \\
V2B~\cite{hui20213d}     & 1.36 M     & 5.57 G  & 39      &70.5 \\
STNet~\cite{stnet}   &1.66 M &3.14 G  &36  &72.1\\
M2Track~\cite{zheng2022beyond} &2.24 M &2.54 G &37  &65.5 \\
\textbf{SyncTrack} & 1.47 M     & \textbf{2.51 G}    & 45    &\textbf{73.3}  \\ \hline
\end{tabular}
\end{table}

\begin{figure*}[h!]
\centering
\includegraphics[width=0.9\linewidth,trim={0cm 0cm 0cm 0cm}]{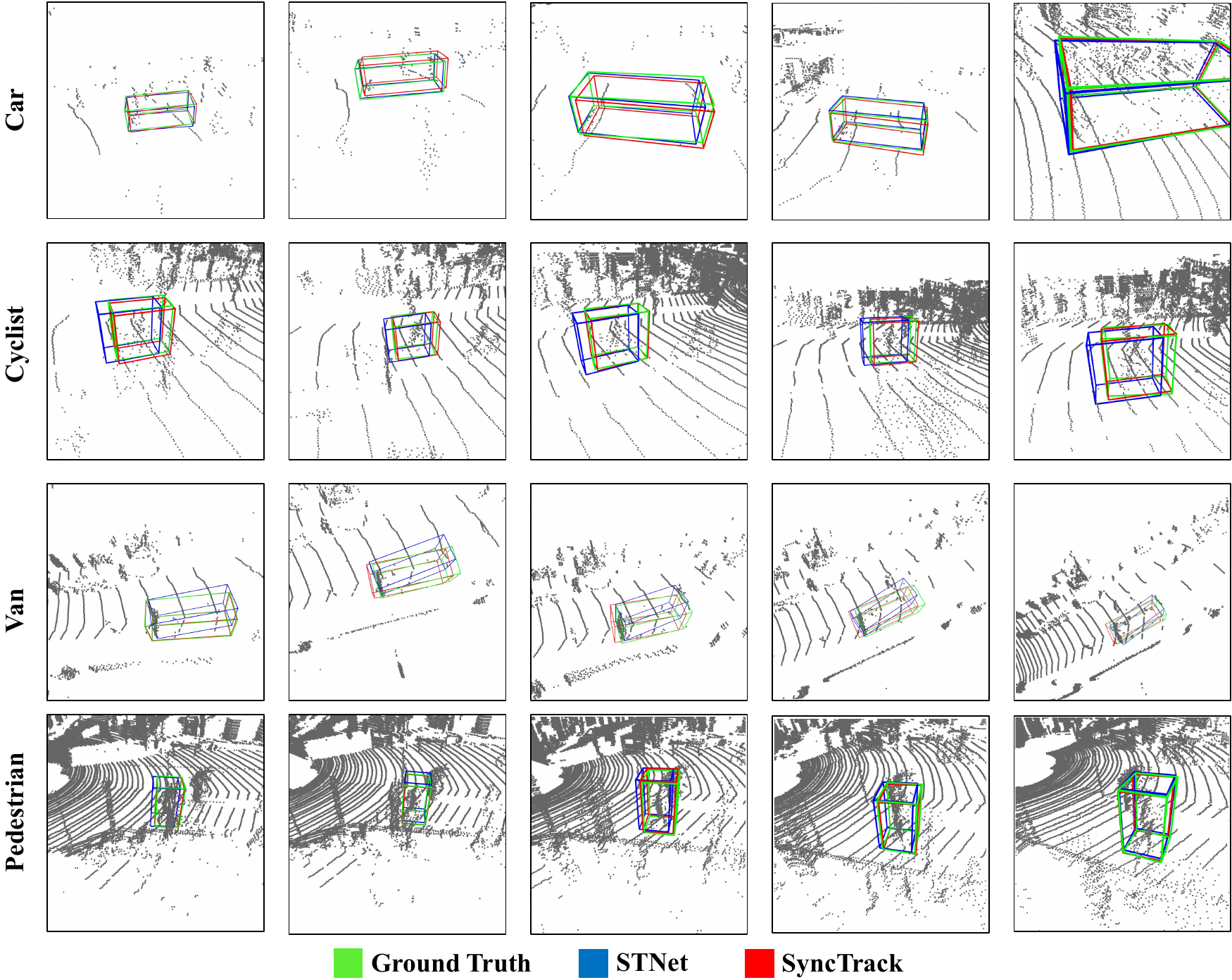}
\caption{\textbf{Qualitative comparison with STNet.} We compare SyncTrack with STNet over four tracking sequences of different categories in KITTI. The predictions of SyncTrack fits the ground-truth boxes better.}
\label{fig:visual}
\end{figure*}

\subsection{Generalization Ability}
To evaluate the generalization ability of SyncTrack, we pre-train the model on the KITTI dataset and test it directly on the nuScenes dataset without fine-tuning. The results are shown in Table~\ref{tab:generalization}. It can be observed that SyncTrack outperforms other methods on the mean results of four categories by a large margin. SyncTrack not only achieves a good balance between inference speed and tracking accuracy, but also generalizes very well to new domains. 

\begin{table}[htbp]
\centering
\caption{Testing results on the nuScenes for generalization ability.}
\renewcommand\arraystretch{1.2}
    \resizebox{\columnwidth}{!}{
    \begin{tabular}{c|c|c|c|c|c}
        \hline
       & Car &Bicycle &Truck &Pedestrian &Mean \\
       \multirow{-2}{*}{Methods} &(15578) &(501) &(3710) &(8019) &(27808)\\ \hline
      SC3D~\cite{giancola2019leveraging} &25.0/27.1 &17.0/18.2 &\textbf{25.7}/\textbf{21.9} &14.2/16.2 &21.8/23.1 \\ 
      P2B~\cite{qi2020p2b} &27.0/29.2 &20.0/26.4 &21.5/16.2 &15.9/22.0 &22.9/25.3 \\
      BAT~\cite{zheng2021box} &22.5/24.1 &17.0/18.8 &19.3/15.8 &17.3/24.5 &20.5/23.0 \\
      V2B~\cite{hui20213d} &31.3/35.1 &\textbf{22.2}/19.1 &21.7/19.1 &17.3/23.4 &25.8/29.0 \\
      STNet~\cite{stnet} &\underline{32.2}/\underline{36.1} &21.2/\textbf{29.2} &22.3/16.8 &\underline{19.1}/\textbf{27.2} &\underline{26.9}/\underline{30.8}\\
      \textbf{SyncTrack} &\textbf{32.8}/\textbf{36.3} &\underline{21.7}/\underline{28.3} &\underline{23.9}/\underline{19.2} &\textbf{19.3}/\underline{27.1} &\textbf{27.5}/\textbf{31.2} \\
      \hline    
    \end{tabular}}
    \label{tab:generalization}
\end{table}

\subsection{Scalability of Backbone~}
We prove that our SyncTrack has good scalability in large-scale datasets like nuScenes. The basic model of SyncTrack has merely one Transformer layer in each stage to ensure real-time performance for tracking. However, the backbone of SyncTrack is scalable in both depth and width. We name the basic model as SyncTrack-Small. Furthermore, the SyncTrack-Mid has three Transformer layers at every stage, with nine layers total. As for the SyncTrack-Large, it doubles the number of feature channels of the SyncTrack-Mid to [256, 128, 64] for every stage. Table~\ref{tab:scale} reveals that performance improves when scalability increases not only in depth (Small \textit{v.s.} Mid), but also in width (Mid \textit{v.s.} Large).

\begin{table}[htbp]
\centering
\small
\caption{The scalability of SyncTrack on nuScenes, (S, M, L represent small, middle, large model size of SyncTrack).}
\label{tab:scale}
\resizebox{\columnwidth}{!}{
\begin{tabular}{c|c|cccc}
\hline
&Scale  & \#Param & FLOPs   &Success &Precision \\ \hline \hline
\multirow{3}*{Bicycle} &S & 1.47 M     &2.51 G    &23.8    &30.4  \\ 
&M & 1.82 M     &2.63 G    & 25.0    &33.6  \\ 
&L & 3.98 M     &5.37 G    & 25.6    &34.1  \\ \hline
\multirow{3}*{Truck} & S & 1.47 M     &2.51 G    & 39.4    &38.6  \\ 
&M & 1.82 M     &2.63 G    & 40.1    &38.8  \\ 
&L & 3.98 M     &5.37 G    & 40.5    &38.9  \\ \hline
\end{tabular}}
\vspace{-10pt}
\end{table}

\subsection{Ablation Studies}
We conduct comprehensive ablations to evaluate the components of SyncTrack.

\noindent\textbf{Compare with Siamese Structure.~}
The synchronized feature extracting and matching mechanism effectively aggregates features and models the relation. To make comparisons, we split the single branch into a Siamese structure based on SyncTrack. A matching network is added to the Siamese backbone for correlating the features. The shape-aware feature learning network in V2B~\cite{hui20213d} and iterative coarse-to-fine correlation network in STNet~\cite{stnet} are chosen as matching networks to compare with our single-branch framework. From Table~\ref{tab:ablation_siamese}, the single-branch structure of SyncTrack is quite significant, and when we split the branch and add a matcher to correlate, the performance drops heavily.

\begin{table}[h!]
    \centering
    \caption{Ablations of proposed Single-Branch and Siamese Structure on \textit{Car} category of KITTI and nuScenes, (matcher1 is from the V2B~\cite{hui20213d} and matcher2 is from the STNet~\cite{stnet}).}
    \Large
    \resizebox{\columnwidth}{!}{
    \begin{tabular}{c|cc|cc}
    \hline
        & \multicolumn{2}{c|}{KITTI}  & \multicolumn{2}{c}{nuScenes} \\ \cline{2-5}
        \multirow{-2}{*}{Structure} & Success & Precision & Success & Precision \\ \hline \hline
        Siamese+matcher1 &70.4 &82.4 &35.3 &36.8 \\
        Siamese+matcher2 &71.8 &83.7 &35.4 &37.1 \\
        Single-Branch &\textbf{73.3} &\textbf{85.0} &\textbf{36.7} &\textbf{38.1} \\ \hline
    \end{tabular}
    }
    \label{tab:ablation_siamese}
\end{table}

\noindent\textbf{Attentive Sampling.~}
In this paper, we integrate attentive sampling into the multi-head Transformers for selecting search region points to aggregate neighborhood features. We ablate such configuration by performing attentive sampling on template tokens and both template and search region tokens, as well as comparing with standard Transformer without attentive sampling (using random sampling and random/FPS sampling) as shown in Table~\ref{tab:ablation_sample}.
It can be concluded that the pattern of performing attentive sampling on search region tokens in Transformers is the best. We hypothesize that sampling template tokens attentively is meaningless as template points are target-centric, and search regions' responses include much noise from the background. Therefore, it i s inefficient to down-sample template tokens based on attentive responses.

\begin{table}[h!]
    \centering
    \small
    \caption{Ablations of attentive sampling in APST on \textit{Car} category of KITTI.}
    \resizebox{\columnwidth}{!}{
\begin{tabular}{c|cc|cc}
\hline
&Template & Search Region   & Success & Precision  \\ \hline
& \XSolidBrush &\XSolidBrush & 70.9   &  82.8              \\
\Checkmark APST  &\Checkmark &\XSolidBrush  &  67.6  &   78.8 \\
 \XSolidBrush random &\XSolidBrush & \Checkmark &\textbf{73.3}  &\textbf{85.0}     \\ 
 &\Checkmark & \Checkmark &69.6  &81.1 \\ \hline  
& \XSolidBrush &\XSolidBrush & 71.1 &82.8              \\
\Checkmark APST  &\Checkmark &\XSolidBrush  &  68.0  &   78.4 \\
 \XSolidBrush FPS &\XSolidBrush & \Checkmark &\textbf{73.2}  &\textbf{85.0}     \\ 
 &\Checkmark & \Checkmark &70.2  &82.7 \\ \hline  
\end{tabular}
}
\vspace{-10pt}
    \label{tab:ablation_sample}
\end{table}

\subsection{Visualization}
In Figure~\ref{fig:visual}, we present visualization results obtained from LiDAR video sequences taken from the KITTI dataset. These visualizations show the motion pattern of objects belonging to four categories wihtin the KITTI dataset. 
Obviously, our SyncTrack excels in accurately tracking the intended target and predicting bounding boxes when compared to STNet~\cite{stnet}. This achievement is primarily attributed to the dynamic and abundant feature interactions that occur between the template and search region seeds in SyncTrack. These interactions enable our algorithm to effectively distinguish the foreground from the background, leading to superior performance.

\section{Conclusion}
In this paper, we propose SyncTrack, a novel single-branch and single-stage framework for 3D LiDAR single object tracking. SyncTrack replaces the conventional Siamese-like backbones with a single-branch one, synchronizing the feature extracting and matching without an additional matching network. Moreover, the Attentive Points-Sampling Transformer is proposed for building the backbone, and sampling search region points attentively rather than randomly. Our SyncTrack achieves good tracking performance in accuracy, efficiency, and scalability. We hope it can help motivate further research on more simple yet efficient 3D trackers.

\textbf{Limitations discussion.}
Compared with motion-centric tracking framework like M2Track~\cite{zheng2022beyond}, we find our SyncTrack achieves limited performance on tiny-sized and slow-moving objects like pedestrian. We attribute it to the global reasoning mechanism of self-attention.
Specifically, the semantic density of small-sized objects' token is much lower, which
hinders effective informative interactions between tokens
when performing self-attention. The fact that transformer-
based method STNet~\cite{stnet} outperforms CNN-based M2track on
all classes except the pedestrian (Table~\ref{tab:kitti}) also supports this hypothesis.

{\small
\bibliographystyle{ieee_fullname}
\bibliography{egbib}
}

\section*{Supplementary Material}
\appendix
\section{Decoder Network and Losses}
With the encoded point-wise features of various scales, a multi-scale feature fusion module is adopted to fuse the search region features only and output the features with the same origin input size, feeding into the decoder part for final predictions.
Following the V2B~\cite{hui20213d}, the features are voxelized as a volumetric representation and 3D convolutions are utilized on the encoded features. 
To ensure the features with high response to the target can be distinguished from all features, max-pooling operation along the \textit{z}-axis is adopted to acquire the BEV feature maps for regression. 
Afterward, layers of 2D convolution blocks (2D convolution, batch normalization and ReLU activation) are leveraged to aggregate the features from dense BEV feature maps, thus the local representations can be captured for the potential target. The decoding process is anchor-free and enjoys the accurate localization due to the perspective of BEV.

Focal loss~\cite{lin2017focal} and L1 loss are leveraged for classification and regression, respectively. 
Following the V2B~\cite{hui20213d}, the 2D target center $(c_x, c_y)$ can be parameterized as $(\frac{x-x_{min}}{v_x}, \frac{y-y_{min}}{v_y})$, where $x_{min}$ and $y_{min}$ are the lower limit of $x$ and $y$ dimension in search area, and $v_x$, $v_y$ are the voxel size in $x-y$ plane. The discrete 2D center is defined by $\hat{c_x}=\lfloor c_x\rfloor$ and $\hat{c_y}=\lfloor c_y\rfloor$. For the pixel $(i,j)$ in the 2D bounding box, if $(i,j)$ is the center of target, then the ground truth classification $\mathcal{G}_{cls}$ is 1, otherwise $\frac{1}{\gamma+1}$, where $\gamma$ is the Euclidean distance between $(i, j)$ and the target center. $\mathcal{G}_{cls}$ equals 0 if a pixel is outside of the bounding box. Based on that, Focal loss is adopted for the classification. For the offset head, the ground truth is $\mathcal{G}_{reg}\in \mathbb{R}^{3\times r\times r}$, and the $r$ is the radius of the object center. The regression target is $[c_x-\hat{c_x}, c_y-\hat{c_y}, \theta]$, where $\theta$ is the rotation angle. Also, the ground truth of $z$-axis $\mathcal{G}_z$ is also considered. Therefore, L1 loss is utilized for both the offset regression and $z$-axis regression. The coefficient is $1,1,2$ for the Focal loss, offset L1 loss and $z$-axis L1 loss, respectively.

\end{document}